\pdfoutput=1
\documentclass[runningheads]{llncs}
\usepackage[T1]{fontenc}
\usepackage{graphicx}
\usepackage{longtable}
\usepackage{booktabs}
\usepackage{array}
\usepackage{bbding}
\usepackage{url}
\usepackage{color}
\usepackage[colorlinks=true,linkcolor=blue,citecolor=blue,urlcolor=blue]{hyperref}
\newcommand{\best}[1]{\underline{#1}}
\newcommand{\ours}[1]{\textbf{#1}}
\newcommand{\second}[1]{#1}

\begin{document}
\title{DirEAG: Dirichlet Evidence Aggregation for Calibrating Verbalized Confidence in Mathematical Reasoning}
\titlerunning{DirEAG for Calibrating Verbalized Confidence}
\author{Haorui Xu\inst{1}\orcidID{0009-0005-4187-0971} \and
Yuzhou Zhu\inst{2}\orcidID{0009-0005-6234-9501} \and
Liyuan Gao\inst{1}\Envelope}
\authorrunning{H. Xu et al.}
\institute{School of Mathematics, Jilin University, Changchun, China\\
\email{hrxu1023@mails.jlu.edu.cn, gaoly@jlu.edu.cn} \and
Leicester International Institute, Dalian University of Technology, Dalian, China\\
\email{1730694701@mail.dlut.edu.cn}}
\maketitle
\begin{abstract}
Reliable confidence estimation is essential for using large language models in mathematical reasoning, but black-box verbalized confidence is difficult to calibrate. When the same problem is queried under multiple confidence-steering prompts, the resulting answer-confidence observations contain useful uncertainty information, yet their scales may shift across steering levels, models, and datasets. Existing black-box uncertainty methods often rely on answer agreement, sample consistency, or entropy, which describe output variation but do not model the numerical meaning of self-reported confidence. Conversely, direct averaging or heuristic aggregation of elicited confidence cannot learn prompt- and task-dependent bias. We propose DirEAG, a Dirichlet Evidence Aggregation method that converts each elicited answer-confidence observation into calibrated soft evidence over generated candidate answers and an additional null state, allowing the model to represent cases where none of the candidates is correct. Experiments on GSM8K, SVAMP, and GSM-Hard with Qwen, Mistral, and Gemma models show that, compared with direct confidence averaging and heuristic confidence-steering aggregation, DirEAG often achieves better calibration while maintaining competitive answer selection. Ablations further reveal that evidence aggregation and final binary calibration address distinct parts of the calibration problem. \emph{Code:} \url{https://github.com/horacehsugithub/DirEAG}.

\end{abstract}
\section{Introduction}
Large language models (LLMs) are increasingly used for mathematical reasoning, where a system is expected not only to produce a final answer, but also to indicate whether that answer should be trusted. Recent prompting and decoding methods have substantially improved arithmetic reasoning performance \cite{wei2022cot,wang2023selfconsistency}. This progress makes confidence estimation increasingly important: unreliable confidence makes it difficult to assess whether an answer should be trusted, while high confidence in an incorrect answer can be actively misleading. Because mathematical answers are often discrete and externally checkable, mathematical reasoning is a particularly suitable testbed for studying whether LLM confidence is aligned with answer-level correctness.

This paper studies confidence estimation in a black-box mathematical reasoning setting. For each problem, an LLM can be prompted to generate a numerical answer and a verbalized confidence score. Because internal probabilities may be unavailable or poorly aligned with answer-level correctness, verbalized confidence provides a readily accessible but biased uncertainty signal \cite{lin2022uncertainty,tian2023justask,xiong2024uncertainty}. Recent work also shows that a model can be queried under different confidence-steering prompts, yielding several answer-confidence observations of the same problem \cite{zhou2025steerconf}. In mathematical reasoning, these observations naturally form a finite set of candidate answers, each supported by one or more reported confidence values.

The resulting technical problem is not simply how to ask an LLM for confidence, but how to convert multiple biased confidence reports into a calibrated probability for the final answer. Existing black-box uncertainty signals often rely on answer agreement, sample consistency, or entropy over sampled outputs \cite{wang2023selfconsistency,lyu2025sampleconsistency,farquhar2024semanticentropy}. These signals are valuable, but they mainly describe variation among outputs and do not directly model the numerical meaning of self-reported confidence. Other methods elicit verbalized confidence or steer the model toward cautious or confident responses, then combine the resulting scores using descriptive statistics or heuristic aggregation rules \cite{tian2023justask,xiong2024uncertainty,zhou2025steerconf}. Thus, existing methods either discard numerical confidence values or use them without learning how model-, prompt-, and task-dependent reporting biases translate into empirical correctness.

The core difficulty is that a verbalized confidence score may contain useful ordinal information while its scale shifts across prompts, models, and datasets. Directly treating self-reported confidence as a probability is therefore too strong, whereas discarding it reduces the problem to answer counting. We argue that multi-prompt confidence estimation should instead use calibration examples to learn how self-reports should influence candidate answers.

We propose \textbf{DirEAG}, a Dirichlet Evidence Aggregation method for calibrating verbalized confidence in mathematical reasoning. DirEAG treats each generated candidate answer as a categorical state and each elicited confidence report as calibrated soft evidence for one of these states. It also includes a null state, which is crucial because aggregation over generated candidates alone cannot represent the case where all generated answers are wrong. The resulting Dirichlet posterior provides a compact probabilistic interface for combining biased black-box observations before applying a final post-hoc binary calibration step to the selected answer confidence.

We evaluate DirEAG on multiple mathematical reasoning benchmarks using Qwen, Mistral, and Gemma models. We compare against vanilla verbalized confidence, mean confidence, self-consistency, answer-entropy confidence, Top-K prompting, and SteerConf-style confidence-steering aggregation. DirEAG improves over direct confidence averaging and heuristic steering aggregation on calibration metrics in most settings, while remaining competitive in answer selection. Internal ablations further separate the effect of evidence aggregation from the correction of the final probability scale.

Our contributions are summarized as follows:
\begin{itemize}
    \item We identify multi-prompt verbalized confidence aggregation as a statistical calibration problem for black-box mathematical reasoning, where multiple biased self-reports must be mapped into an answer-level probability.
    \item We propose DirEAG, a learned aggregation model with few trainable parameters for calibrating self-reported confidence, weighting confidence-elicitation levels, and accumulating evidence in a Dirichlet model over generated candidate answers and an explicit null state.
    \item We provide experiments and ablations showing that confidence evidence and final binary calibration play complementary roles: the former affects candidate scoring and selection, while the latter corrects the selected-answer probability scale.
\end{itemize}

\section{Related Work}

\subsection{Verbalized Confidence in LLMs}

Verbalized confidence estimates uncertainty by asking a language model to state how confident it is in its own answer. Lin et al. \cite{lin2022uncertainty} showed that models can be trained to express uncertainty in words. Tian et al. \cite{tian2023justask} found that verbalized confidence from RLHF models can be better calibrated than conditional token probabilities, while Xiong et al. \cite{xiong2024uncertainty} showed that LLMs still often exhibit overconfidence under black-box confidence elicitation. Prompt design also matters: direct confidence prompts, Top-K prompts, and chain-of-thought confidence prompts can affect reported confidence \cite{tian2023justask,xiong2024uncertainty}; epistemic markers in prompts can also change model behavior \cite{zhou2023greyarea}. SteerConf \cite{zhou2025steerconf} further uses confidence-steering prompts to obtain multiple answer-confidence observations from the same model. These studies suggest that verbalized confidence contains useful uncertainty information, but should not be treated as a calibrated probability by default.

\subsection{Confidence Calibration for LLMs}

Confidence calibration for LLMs asks whether stated confidence matches empirical correctness. Prior work approaches this problem through confidence elicitation, probability baselines, and black-box aggregation. Tian et al. \cite{tian2023justask} compare verbalized confidence with probability-based alternatives, while Xiong et al. \cite{xiong2024uncertainty} show that LLMs can remain overconfident under explicit uncertainty prompts. SteerConf \cite{zhou2025steerconf} further exploits confidence steerability by querying the same model under multiple confidence prompts and aggregating the resulting observations.

Recent work has also explored stronger forms of verbalized-confidence calibration. Wang and Stengel-Eskin \cite{wang2026dinco} propose DINCO, which normalizes verbalized confidence against self-generated distractors to reduce suggestibility-driven overconfidence. Li et al. \cite{li2025conftuner} propose ConfTuner, a training-based approach that teaches LLMs to express confidence verbally using a tokenized Brier-score objective. These methods are complementary to our setting: DINCO introduces additional distractor-based validation, while ConfTuner modifies the model through fine-tuning. In contrast, DirEAG focuses on post-hoc aggregation of already elicited answer-confidence observations from a fixed black-box model.

More specifically, our work focuses on the aggregation step after answer-confidence pairs have been obtained. This makes the calibration problem depend not only on how confidence is elicited, but also on how repeated and conflicting observations are combined.

\subsection{Multi-observation Uncertainty Estimation}

Multiple generations provide another black-box route to uncertainty estimation. Chain-of-thought prompting \cite{wei2022cot} and self-consistency \cite{wang2023selfconsistency} show that sampled reasoning paths can improve mathematical reasoning, while answer agreement also gives a simple uncertainty signal. Sample consistency explicitly uses the distribution of sampled generations for calibration \cite{lyu2025sampleconsistency}. SelfCheckGPT \cite{manakul2023selfcheckgpt} and semantic entropy \cite{farquhar2024semanticentropy} similarly estimate uncertainty from agreement or semantic diversity among multiple outputs.

DirEAG is related to these multi-observation methods, but the observations have a different structure: each confidence-steered output contains both a candidate answer and a reported confidence score. For numerical reasoning, this naturally yields a finite candidate set. Our method uses this structure directly instead of reducing the observations to agreement frequency alone.

\section{Method}

We study confidence estimation for mathematical reasoning when a language model is queried under multiple confidence-steering prompts. Given the resulting answer-confidence observations, our goal is to estimate a probability distribution over generated candidates with an explicit statistical interpretation.

\begin{figure}[!ht]
\centering
\includegraphics[width=\textwidth,trim=4pt 85pt 4pt 75pt,clip]{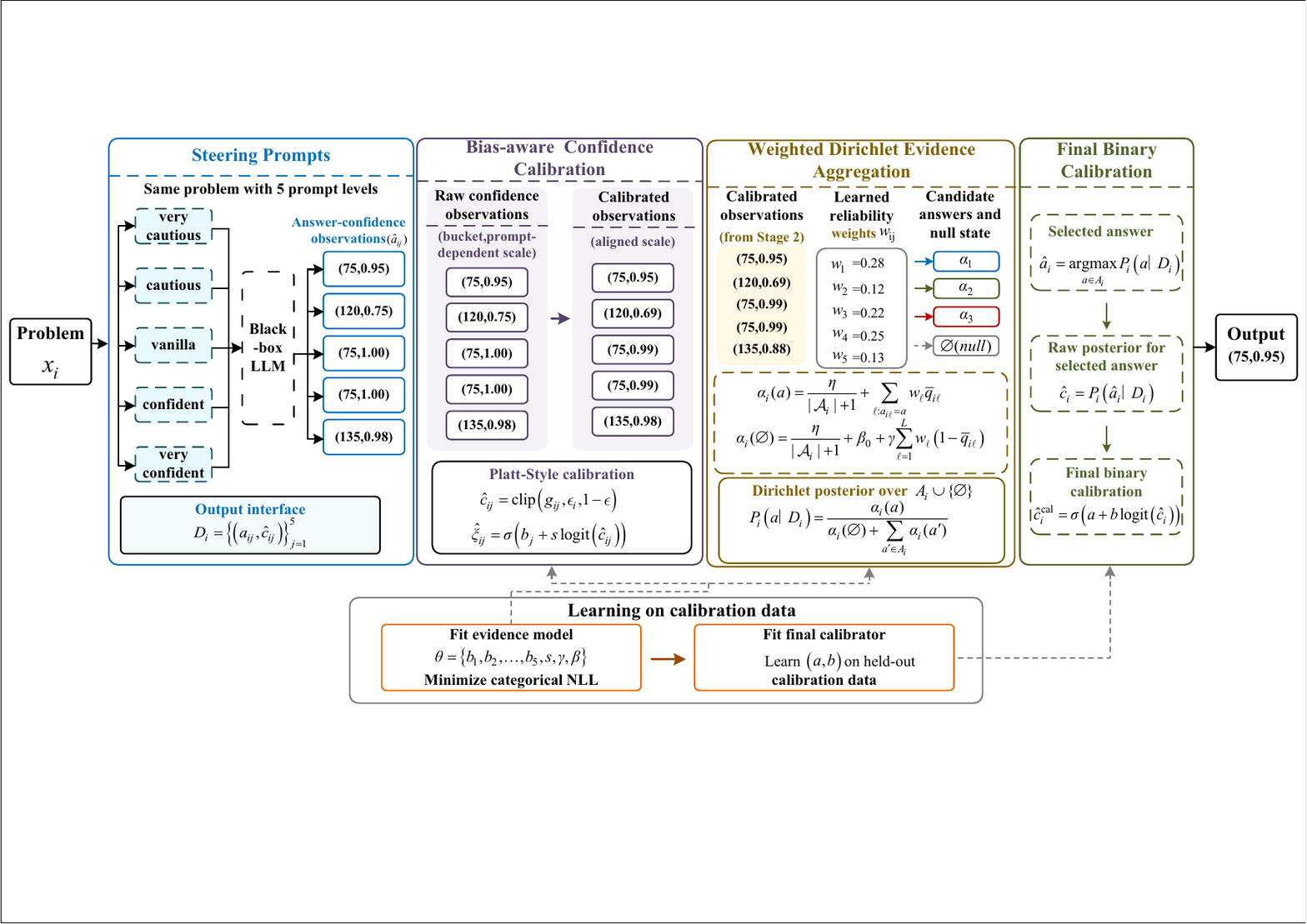}
\caption{Overview of DirEAG. Multiple answer-confidence observations are converted into calibrated soft evidence, aggregated over candidate answers and a null state, and finally calibrated as a selected-answer correctness probability.}
\label{fig:method-overview}
\end{figure}

Figure~\ref{fig:method-overview} summarizes the overall pipeline. The model is queried with five confidence-steering prompts, producing answer-confidence observations on biased scales. DirEAG calibrates these reports, aggregates the resulting soft evidence in an augmented candidate space, and maps the selected posterior mass to a calibrated correctness probability.

\subsection{Biased Soft Evidence from Confidence Elicitation}

Let $D_i=\{(a_{i\ell},q_{i\ell})\}_{\ell=1}^L$ denote the observed outputs for problem $x_i$. We view each pair $(a_{i\ell},q_{i\ell})$ as soft evidence in favor of the generated answer $a_{i\ell}$. Following the five-level confidence elicitation design of SteerConf, we query the model under confidence-steering prompts that induce different attitudes toward confidence reporting, ranging from cautious to highly confident. We adopt this elicitation component because it provides a standardized and model-agnostic way to obtain multiple answer-confidence observations of the same problem. 

Because the confidence value $q_{i\ell}$ is an elicited self-report, its scale may not be directly comparable across steering levels. A more cautious prompt and a more confident prompt may produce scores on different biased scales, even when their answers are generated by the same underlying model.

To account for this prompt-dependent scale bias, we apply a monotonic calibration transform with few trainable parameters to each raw confidence score. The transform follows the logit-sigmoid form of Platt-style calibration \cite{platt1999probabilistic,guo2017calibration}, which is suitable here because verbalized confidence is an ordered scalar signal whose scale can shift across elicitation levels. Since the logit function is undefined at the boundary values $0$ and $1$, we first clip the raw confidence score into the open interval $(0,1)$:
\begin{equation}
q_{i\ell}^{\epsilon}
=
\mathrm{clip}(q_{i\ell},\epsilon,1-\epsilon),
\qquad 0<\epsilon<\frac{1}{2}.
\end{equation}
The calibrated confidence score is then defined as
\begin{equation}
\tilde q_{i\ell}
=
\sigma\left(b_\ell+s\,\mathrm{logit}(q_{i\ell}^{\epsilon})\right),
\qquad s=\mathrm{softplus}(r)>0,
\end{equation}
where $b_\ell$ is a steering-level-specific bias, $s$ is a positive shared slope, and $\sigma(\cdot)$ denotes the logistic sigmoid. The positive parameterization ensures that the transform is monotone non-decreasing in the raw reported confidence, so it can correct the probability scale without reversing within-level confidence ordering. The shared slope keeps the calibration low-dimensional, while $b_\ell$ captures prompt-dependent bias.

\subsection{Dirichlet Evidence Aggregation over an Augmented Candidate Space}

In mathematical reasoning, the final prediction is typically a discrete numerical answer. Multiple prompted outputs therefore induce a finite candidate set:
\begin{equation}
\mathcal A_i=\{a_{i\ell}:\ell=1,\dots,L\}.
\end{equation}
The aggregation problem can then be expressed as inference over which candidate answer is correct. This is a categorical uncertainty problem rather than a single binary confidence problem.

We also introduce a null state $\emptyset$, representing the event that the gold answer is absent from the generated candidate set. This state is necessary because aggregation over generated candidates alone cannot represent candidate-set failure.
\begin{equation}
\mathcal S_i=\mathcal A_i\cup\{\emptyset\}.
\end{equation}

We use the Dirichlet distribution as an evidence parameterization over the question-specific candidate state space. It supports a variable number of generated answer states, represents accumulated non-negative evidence before normalization, and places the null state on the same categorical simplex as generated candidates.

We use its posterior mean as the candidate-level support distribution. For each candidate answer $a\in\mathcal A_i$, we define
\begin{equation}
\alpha_i(a)
=
\frac{\eta}{|\mathcal A_i|+1}
+
\sum_{\ell:a_{i\ell}=a} w_\ell \tilde q_{i\ell},
\end{equation}
where $\eta>0$ is the total symmetric prior strength and $w_\ell\geq 0$ is the learned reliability weight of steering level $\ell$. Thus, repeated answers accumulate more concentration only when supported by reliable prompts with calibrated high confidence.

The null state receives prior mass and uncertainty evidence:
\begin{equation}
\alpha_i(\emptyset)
=
\frac{\eta}{|\mathcal A_i|+1}
+
\beta_0
+
\gamma\sum_{\ell=1}^{L} w_\ell(1-\tilde q_{i\ell}),
\end{equation}
where $\beta_0\geq 0$ is a base concentration for the null state and $\gamma\geq 0$ controls how much low calibrated confidence contributes to candidate-missing uncertainty. This lets candidate-set failure be incorporated into the same normalized state space, lowering selected-answer support when elicited observations are weak.

The posterior mean over states is
\begin{equation}
P_i(a\mid D_i)
=
\frac{\alpha_i(a)}
{\alpha_i(\emptyset)+\sum_{b\in\mathcal A_i}\alpha_i(b)}.
\end{equation}
The final answer is selected from generated candidates by maximum posterior probability:
\begin{equation}
\hat a_i=\arg\max_{a\in\mathcal A_i}P_i(a\mid D_i),
\end{equation}
and its raw aggregated confidence is
\begin{equation}
\hat c_i=P_i(\hat a_i\mid D_i).
\end{equation}

\subsection{Learning and Binary Calibration}

Let $y_i$ be the gold answer. The target state is the gold answer if it appears in the candidate set, and the null state otherwise:
\begin{equation}
t_i=
\left\{
\begin{array}{ll}
	y_i, & y_i\in\mathcal A_i,\\
	\emptyset, & y_i\notin\mathcal A_i.
\end{array}
\right.
\end{equation}
We estimate
\begin{equation}
\theta=\{w_\ell,b_\ell,s,\eta,\beta_0,\gamma\}
\end{equation}
by minimizing the categorical negative log-likelihood. Nonnegative parameters, including $w_\ell$, $s$, $\eta$, $\beta_0$, and $\gamma$, are represented with positive parameterizations such as softplus:
\begin{equation}
\mathcal L(\theta)
=
-\sum_i \log P_i(t_i\mid D_i;\theta)
+
\lambda\|\theta\|_2^2.
\end{equation}
All parameters are shared across examples within the same calibration setting.

The Dirichlet evidence model produces a posterior distribution over generated candidates and the null state. The selected posterior mass $\hat c_i$ is a within-instance measure of relative support: it indicates how much the observed answer-confidence pairs support the selected candidate compared with other generated candidates and the null state. Calibration metrics, however, require an across-instance binary reliability estimate: among predictions assigned a given confidence value, what fraction of selected answers are actually correct? These quantities need not coincide, because the Dirichlet model is only an approximate evidence model and the max-selection step turns a candidate-level distribution into a binary correctness event.

Therefore, we apply a second Platt-style calibration \cite{platt1999probabilistic,guo2017calibration} to map the selected posterior mass to an empirical correctness probability. This two-parameter mapping directly corrects the final probability scale on a held-out calibration split:
\begin{equation}
\hat c_i^{\mathrm{cal}}
=
\sigma\left(a+b\,\mathrm{logit}(\hat c_i)\right).
\end{equation}
The Platt parameters are fitted on this held-out split. This step does not change the evidence aggregation mechanism; it converts the aggregated score into the binary probability required by calibration metrics such as ECE and Brier score.

\section{Experiments}

\subsection{Experimental Setup}

\paragraph{Tasks and Metrics.}
The experiments measure both answer selection and confidence quality. We report accuracy (Acc), expected calibration error (ECE), Brier score, AUROC, PR-P, and PR-N. Accuracy measures whether the selected answer is correct, while ECE and Brier score evaluate probabilistic calibration. AUROC measures the ranking quality of confidence scores for separating correct and incorrect predictions. PR-P treats correct predictions as the positive class, whereas PR-N treats incorrect predictions as the positive class. ECE is computed with 10 equal-width bins, and the released code includes the exact metric implementation used for all reported results.

\paragraph{LLM Models.}
We run the comparison on three instruction-tuned open-weight models: Qwen2.5-7B-Instruct \cite{qwen2024qwen25}, Mistral-7B-Instruct-v0.3 \cite{jiang2023mistral}, and Gemma-2-9B-it \cite{gemma2024gemma2}. Together, these models provide a moderate-scale testbed for assessing whether the proposed aggregation method behaves consistently across model families rather than on a single backbone.

\paragraph{Datasets.}
We evaluate on three mathematical reasoning benchmarks: GSM8K \cite{cobbe2021gsm8k}, SVAMP \cite{patel2021svamp}, and GSM-Hard \cite{gao2023pal}. This choice keeps the study focused on numerical reasoning, where final-answer correctness can be evaluated by extracting the predicted value and comparing it with the gold answer. GSM8K uses the official train/test split, whereas SVAMP and GSM-Hard use 5-fold cross-fitting. In all cases, evaluated examples are excluded from both DirEAG parameter fitting and final Platt calibration. Optimization settings are fixed across model--dataset pairs, with full details released in the code.

\paragraph{Baselines.}
We align methods by using five answer-confidence observations whenever possible: DirEAG, mean confidence, and SteerConf share the same five answer-confidence observations, self-consistency and answer-entropy confidence use five vanilla samples, and Top-$K$ is included as a direct single-response baseline that elicits five candidates at once. Vanilla uses the neutral confidence prompt. SteerConf applies the original confidence-steering aggregation rule. DirEAG denotes our Dirichlet evidence aggregation method with few trainable parameters for confidence calibration and final binary calibration. Published baselines are cited directly in Table~\ref{tab:main-results}.

\subsection{Main Results}
	
	\begingroup
	\footnotesize
	\setlength{\tabcolsep}{1.8pt}
	\begin{longtable}{@{}l>{\raggedright\arraybackslash}p{2.45cm}rrrrrr@{}}
		\caption{Main comparison across three models and three datasets. $\uparrow/\downarrow$ indicate preferred direction; \best{underline} marks the best value within each model--dataset block, and \ours{bold} marks DirEAG.}\label{tab:main-results}\\
		\toprule
		Dataset & Method & Acc$\uparrow$ & ECE$\downarrow$ & Brier$\downarrow$ & AUROC$\uparrow$ & PR-P$\uparrow$ & PR-N$\uparrow$\\
		\midrule
		\endfirsthead
		\toprule
		Dataset & Method & Acc$\uparrow$ & ECE$\downarrow$ & Brier$\downarrow$ & AUROC$\uparrow$ & PR-P$\uparrow$ & PR-N$\uparrow$\\
		\midrule
		\endhead
		\bottomrule
		\endfoot
		\multicolumn{8}{c}{\textbf{Model: Qwen2.5-7B}}\\
		\midrule
		GSM8K & Vanilla & 0.7582 & 0.2150 & 0.2174 & 0.7077 & 0.8479 & 0.4161\\
		 & Self-cons.~\cite{wang2023selfconsistency} & \best{0.9265} & \best{0.0170} & \best{0.0486} & 0.8542 & \second{0.9784} & 0.4244\\
		 & Ans. entropy~\cite{farquhar2024semanticentropy} & \best{0.9265} & 0.0384 & \second{0.0524} & 0.8538 & \best{0.9786} & 0.4241\\
		 & Mean conf. & \second{0.8992} & 0.1049 & 0.0840 & 0.8461 & 0.9723 & 0.5098\\
		 & SteerConf~\cite{zhou2025steerconf} & 0.7930 & 0.0473 & 0.1075 & \best{0.8813} & 0.9602 & \second{0.6767}\\
		 & Top-$K$~\cite{xiong2024uncertainty} & 0.5853 & 0.3021 & 0.2719 & 0.8552 & 0.8396 & \best{0.8051}\\
		 & \textbf{DirEAG} & \ours{0.8908} & \ours{\second{0.0280}} & \ours{0.0736} & \ours{\second{0.8621}} & \ours{0.9744} & \ours{0.5136}\\
		\addlinespace[1pt]
		SVAMP & Vanilla & 0.7780 & 0.1919 & 0.2007 & 0.6313 & 0.8301 & 0.3295\\
		 & Self-cons.~\cite{wang2023selfconsistency} & \best{0.8880} & 0.0754 & \second{0.0862} & 0.7006 & 0.9303 & 0.3502\\
		 & Ans. entropy~\cite{farquhar2024semanticentropy} & \best{0.8880} & 0.0761 & \best{0.0848} & 0.7008 & 0.9304 & 0.3646\\
		 & Mean conf. & 0.8440 & 0.0477 & 0.0928 & \second{0.8275} & \best{0.9462} & \second{0.5824}\\
		 & SteerConf~\cite{zhou2025steerconf} & 0.7950 & \second{0.0315} & 0.1024 & \best{0.8581} & \second{0.9397} & \best{0.6869}\\
		 & Top-$K$~\cite{xiong2024uncertainty} & 0.8310 & 0.1434 & 0.1450 & 0.6420 & 0.8740 & 0.3549\\
		 & \textbf{DirEAG} & \ours{\second{0.8510}} & \ours{\best{0.0176}} & \ours{0.0925} & \ours{0.8125} & \ours{0.9383} & \ours{0.5375}\\
		\addlinespace[1pt]
		GSM-Hard & Vanilla & 0.3518 & 0.5949 & 0.5792 & 0.6657 & 0.4618 & 0.7518\\
		 & Self-cons.~\cite{wang2023selfconsistency} & \best{0.5588} & 0.2129 & 0.2129 & 0.8105 & 0.7866 & 0.7554\\
		 & Ans. entropy~\cite{farquhar2024semanticentropy} & \best{0.5588} & 0.1442 & 0.1910 & 0.8111 & \second{0.7891} & 0.7632\\
		 & Mean conf. & \second{0.4564} & 0.1326 & 0.1958 & 0.8058 & 0.7544 & 0.8131\\
		 & SteerConf~\cite{zhou2025steerconf} & 0.3836 & \second{0.1090} & \best{0.1592} & \best{0.8577} & 0.7853 & \best{0.8941}\\
		 & Top-$K$~\cite{xiong2024uncertainty} & 0.3789 & 0.5513 & 0.5254 & 0.7190 & 0.5259 & 0.7860\\
		 & \textbf{DirEAG} & \ours{0.4458} & \ours{\best{0.0556}} & \ours{\second{0.1602}} & \ours{\second{0.8440}} & \ours{\best{0.7951}} & \ours{\second{0.8675}}\\
		\midrule
		\multicolumn{8}{c}{\textbf{Model: Mistral-7B}}\\
		\midrule
		GSM8K & Vanilla & 0.5258 & 0.4510 & 0.4501 & 0.5419 & 0.5477 & 0.5158\\
		 & Self-cons.~\cite{wang2023selfconsistency} & 0.5421 & 0.0650 & 0.1951 & 0.7723 & 0.7377 & 0.7227\\
		 & Ans. entropy~\cite{farquhar2024semanticentropy} & 0.5421 & 0.1500 & 0.2140 & 0.7583 & 0.7430 & 0.7135\\
		 & Mean conf. & \best{0.6346} & \second{0.0589} & \second{0.1504} & \second{0.8489} & \second{0.8949} & 0.7524\\
		 & SteerConf~\cite{zhou2025steerconf} & 0.5102 & 0.0837 & 0.1647 & 0.8468 & 0.8481 & \second{0.8189}\\
		 & Top-$K$~\cite{xiong2024uncertainty} & 0.0728 & 0.8176 & 0.7615 & 0.5517 & 0.0850 & \best{0.9353}\\
		 & \textbf{DirEAG} & \ours{\second{0.6308}} & \ours{\best{0.0585}} & \ours{\best{0.1486}} & \ours{\best{0.8521}} & \ours{\best{0.8988}} & \ours{0.7634}\\
		\addlinespace[1pt]
		SVAMP & Vanilla & 0.6650 & 0.3209 & 0.3203 & 0.5419 & 0.6844 & 0.3782\\
		 & Self-cons.~\cite{wang2023selfconsistency} & 0.5910 & 0.1968 & 0.2295 & 0.7523 & 0.7599 & 0.6572\\
		 & Ans. entropy~\cite{farquhar2024semanticentropy} & 0.5910 & 0.1650 & 0.2157 & 0.7561 & 0.7649 & 0.6762\\
		 & Mean conf. & \best{0.7430} & \second{0.0570} & \best{0.1292} & \second{0.8389} & \best{0.9155} & 0.6857\\
		 & SteerConf~\cite{zhou2025steerconf} & 0.6460 & 0.0789 & 0.1487 & \best{0.8516} & 0.8940 & \best{0.7551}\\
		 & Top-$K$~\cite{xiong2024uncertainty} & 0.5290 & 0.4158 & 0.4314 & 0.5451 & 0.5615 & 0.4907\\
		 & \textbf{DirEAG} & \ours{\second{0.7380}} & \ours{\best{0.0403}} & \ours{\second{0.1304}} & \ours{0.8306} & \ours{\second{0.9135}} & \ours{\second{0.6860}}\\
		\addlinespace[1pt]
		GSM-Hard & Vanilla & 0.1849 & 0.7678 & 0.7655 & 0.5438 & 0.1992 & 0.8312\\
		 & Self-cons.~\cite{wang2023selfconsistency} & 0.2356 & 0.1731 & 0.1614 & 0.8234 & 0.5461 & 0.9158\\
		 & Ans. entropy~\cite{farquhar2024semanticentropy} & 0.2356 & \second{0.0947} & 0.1618 & 0.7636 & 0.5226 & 0.8940\\
		 & Mean conf. & \best{0.2509} & 0.1738 & 0.1585 & 0.8507 & \second{0.6365} & 0.9364\\
		 & SteerConf~\cite{zhou2025steerconf} & 0.1827 & 0.1886 & \second{0.1411} & \best{0.8517} & 0.5817 & \second{0.9507}\\
		 & Top-$K$~\cite{xiong2024uncertainty} & 0.0221 & 0.8963 & 0.8594 & 0.4624 & 0.0225 & \best{0.9741}\\
		 & \textbf{DirEAG} & \ours{\second{0.2502}} & \ours{\best{0.0387}} & \ours{\best{0.1245}} & \ours{\second{0.8513}} & \ours{\best{0.6547}} & \ours{0.9337}\\
		\midrule
		\multicolumn{8}{c}{\textbf{Model: Gemma-2-9B-it}}\\
		\midrule
		GSM8K & Vanilla & 0.8543 & 0.1387 & 0.1369 & 0.5867 & 0.8765 & 0.2821\\
		 & Self-cons.~\cite{wang2023selfconsistency} & \best{0.8952} & 0.0503 & \best{0.0717} & 0.7923 & 0.9548 & 0.4135\\
		 & Ans. entropy~\cite{farquhar2024semanticentropy} & \best{0.8952} & 0.0487 & \second{0.0722} & 0.7918 & 0.9551 & 0.4087\\
		 & Mean conf. & 0.8832 & \second{0.0383} & 0.0754 & 0.8269 & 0.9601 & 0.5203\\
		 & SteerConf~\cite{zhou2025steerconf} & 0.8385 & 0.1049 & 0.0959 & \best{0.8806} & \second{0.9654} & \second{0.6484}\\
		 & Top-$K$~\cite{xiong2024uncertainty} & 0.7437 & 0.1726 & 0.1520 & 0.7735 & 0.8655 & \best{0.6542}\\
		 & \textbf{DirEAG} & \ours{\second{0.8840}} & \ours{\best{0.0292}} & \ours{0.0811} & \ours{\second{0.8342}} & \ours{\best{0.9660}} & \ours{0.4656}\\
		\addlinespace[1pt]
		SVAMP & Vanilla & 0.8060 & 0.1781 & 0.1775 & 0.5627 & 0.8261 & 0.2924\\
		 & Self-cons.~\cite{wang2023selfconsistency} & \best{0.8485} & \second{0.0953} & \second{0.1113} & 0.6978 & \best{0.9032} & 0.4154\\
		 & Ans. entropy~\cite{farquhar2024semanticentropy} & \best{0.8485} & 0.0967 & 0.1133 & 0.6957 & \best{0.9032} & 0.3942\\
		 & Mean conf. & 0.8280 & 0.1092 & 0.1258 & 0.7234 & \second{0.8996} & 0.4535\\
		 & SteerConf~\cite{zhou2025steerconf} & 0.8070 & 0.0954 & 0.1312 & \best{0.7486} & 0.8950 & \best{0.5636}\\
		 & Top-$K$~\cite{xiong2024uncertainty} & \second{0.8347} & 0.1472 & 0.1431 & 0.6107 & 0.8664 & 0.3460\\
		 & \textbf{DirEAG} & \ours{0.8230} & \ours{\best{0.0418}} & \ours{\best{0.1112}} & \ours{\second{0.7293}} & \ours{0.8908} & \ours{\second{0.5335}}\\
		\addlinespace[1pt]
		GSM-Hard & Vanilla & 0.4655 & 0.5008 & 0.4964 & 0.5880 & 0.5148 & 0.6011\\
		 & Self-cons.~\cite{wang2023selfconsistency} & \best{0.5133} & 0.2821 & 0.2580 & 0.7943 & 0.7353 & 0.7614\\
		 & Ans. entropy~\cite{farquhar2024semanticentropy} & \best{0.5133} & 0.2461 & 0.2375 & 0.7995 & 0.7401 & 0.7820\\
		 & Mean conf. & 0.5004 & 0.2849 & 0.2685 & 0.7888 & 0.7318 & 0.7849\\
		 & SteerConf~\cite{zhou2025steerconf} & 0.4731 & \second{0.0844} & \second{0.1715} & \best{0.8394} & \second{0.7999} & \best{0.8399}\\
		 & Top-$K$~\cite{xiong2024uncertainty} & 0.4581 & 0.4872 & 0.4722 & 0.6685 & 0.5613 & 0.6958\\
		 & \textbf{DirEAG} & \ours{\second{0.5042}} & \ours{\best{0.0432}} & \ours{\best{0.1679}} & \ours{\second{0.8274}} & \ours{\best{0.8010}} & \ours{\second{0.8130}}\\
	\end{longtable}
	\endgroup

Vanilla and mean confidence serve as simple verbalized-confidence baselines, while self-consistency and answer-entropy confidence test whether answer agreement alone provides a sufficient uncertainty signal. Across models and datasets, agreement-based baselines remain strong for answer selection and ranking, confirming that repeated answer identity is informative in mathematical reasoning. Top-$K$ prompting is also occasionally competitive in answer accuracy because it directly elicits multiple candidate answers, but its raw confidence can remain high when the selected answer is wrong. Overall, these results show that multi-output structure is useful, but directly using agreement or reported scores does not consistently yield calibrated probabilities.

Compared with direct confidence averaging and SteerConf-style aggregation, DirEAG achieves lower ECE in most settings and often improves answer selection, indicating that confidence-steered reports benefit from learned statistical aggregation.

The ranking-oriented metrics reveal a complementary aspect of confidence quality. AUROC and PR-N depend on whether low-confidence predictions coincide with actual failures, whereas ECE and Brier score evaluate whether the reported probabilities are numerically meaningful. Entropy-based and confidence-steering baselines can therefore remain competitive on failure ranking when answer disagreement is a strong error signal, even if their probability scale is less well calibrated. This contrast suggests that verbalized-confidence aggregation should be assessed along both axes: probabilistic calibration and the prioritization of likely errors.

\subsection{Internal Ablation}

Because DirEAG includes a final binary calibration step, the main comparison should be interpreted as an end-to-end pipeline comparison rather than as an isolated test of the aggregation rule alone. We use the internal ablation in Table~\ref{tab:direag-ablation} to separate the effects of answer counting, confidence evidence, and final binary calibration under the same held-out fitting protocol. Each variant answers a specific question. \emph{Count-only + Cal.} asks whether answer frequency plus final binary calibration is already sufficient. \emph{+ Level Reliability} tests whether different confidence-steering prompts have stable reliability differences. \emph{+ Conf. evidence} asks whether reported confidence can change candidate scoring and selection; it removes final binary calibration to expose the raw aggregated scale. \emph{Full DirEAG} tests whether final binary calibration can turn the selected posterior mass into a reliable correctness probability after evidence aggregation.

\begingroup
\footnotesize
\setlength{\tabcolsep}{1.8pt}
\begin{longtable}{@{}l>{\scriptsize\raggedright\arraybackslash}p{2.45cm}rrrrrr@{}}
\caption{Internal ablation of DirEAG components. $\uparrow/\downarrow$ indicate preferred direction; \best{underline} marks the best value within each block, and \ours{bold} marks Full DirEAG.}\label{tab:direag-ablation}\\
\toprule
Dataset & Variant & Acc$\uparrow$ & ECE$\downarrow$ & Brier$\downarrow$ & AUROC$\uparrow$ & PR-P$\uparrow$ & PR-N$\uparrow$\\
\midrule
\endfirsthead
\toprule
Dataset & Variant & Acc$\uparrow$ & ECE$\downarrow$ & Brier$\downarrow$ & AUROC$\uparrow$ & PR-P$\uparrow$ & PR-N$\uparrow$\\
\midrule
\endhead
\bottomrule
\endfoot
\multicolumn{8}{c}{\textbf{Model: Qwen2.5-7B}}\\
\midrule
GSM8K & Count-only + Cal. & \second{0.8863} & \second{0.0366} & \second{0.0751} & \second{0.8572} & \second{0.9710} & \best{0.5607}\\
 & + Level Reliability & \second{0.8863} & 0.0393 & 0.0755 & \second{0.8572} & \second{0.9710} & \second{0.5606}\\
 & + Conf. evidence & \best{0.8908} & 0.4276 & 0.2619 & \best{0.8621} & \best{0.9744} & 0.5136\\
 & \textbf{Full DirEAG} & \ours{\best{0.8908}} & \ours{\best{0.0280}} & \ours{\best{0.0736}} & \ours{\best{0.8621}} & \ours{\best{0.9744}} & \ours{0.5136}\\
\addlinespace[1pt]
SVAMP & Count-only + Cal. & \second{0.8430} & \second{0.0154} & \second{0.0938} & \best{0.8162} & \best{0.9413} & \best{0.5642}\\
 & + Level Reliability & \second{0.8430} & \best{0.0153} & 0.0939 & \second{0.8161} & \best{0.9413} & \second{0.5632}\\
 & + Conf. evidence & \best{0.8510} & 0.0711 & 0.0966 & 0.8128 & 0.9381 & 0.5484\\
 & \textbf{Full DirEAG} & \ours{\best{0.8510}} & \ours{0.0176} & \ours{\best{0.0925}} & \ours{0.8125} & \ours{\second{0.9383}} & \ours{0.5375}\\
\addlinespace[1pt]
GSM-Hard & Count-only + Cal. & \second{0.4390} & \best{0.0296} & \second{0.1646} & 0.8303 & 0.7561 & 0.8567\\
 & + Level Reliability & \second{0.4390} & \second{0.0298} & \second{0.1646} & 0.8303 & 0.7561 & 0.8567\\
 & + Conf. evidence & \best{0.4458} & 0.1286 & 0.1764 & \best{0.8443} & \second{0.7946} & \second{0.8672}\\
 & \textbf{Full DirEAG} & \ours{\best{0.4458}} & \ours{0.0556} & \ours{\best{0.1602}} & \ours{\second{0.8440}} & \ours{\best{0.7951}} & \ours{\best{0.8675}}\\
\midrule
\multicolumn{8}{c}{\textbf{Model: Mistral-7B}}\\
\midrule
GSM8K & Count-only + Cal. & \second{0.6209} & \best{0.0407} & \best{0.1441} & \best{0.8558} & \second{0.8953} & \best{0.7711}\\
 & + Level Reliability & \second{0.6209} & \second{0.0417} & \best{0.1441} & \best{0.8558} & \second{0.8953} & \best{0.7711}\\
 & + Conf. evidence & \best{0.6308} & 0.2928 & 0.2524 & \second{0.8521} & \best{0.8988} & \second{0.7634}\\
 & \textbf{Full DirEAG} & \ours{\best{0.6308}} & \ours{0.0585} & \ours{\second{0.1486}} & \ours{\second{0.8521}} & \ours{\best{0.8988}} & \ours{\second{0.7634}}\\
\addlinespace[1pt]
SVAMP & Count-only + Cal. & \best{0.7380} & 0.0491 & \best{0.1302} & \second{0.8317} & 0.9112 & 0.6921\\
 & + Level Reliability & \best{0.7380} & \second{0.0488} & \best{0.1302} & 0.8316 & 0.9112 & \second{0.6923}\\
 & + Conf. evidence & \best{0.7380} & 0.3453 & 0.2633 & \best{0.8321} & \best{0.9144} & \best{0.6925}\\
 & \textbf{Full DirEAG} & \ours{\best{0.7380}} & \ours{\best{0.0403}} & \ours{\second{0.1304}} & \ours{0.8306} & \ours{\second{0.9135}} & \ours{0.6860}\\
\addlinespace[1pt]
GSM-Hard & Count-only + Cal. & \second{0.2494} & 0.0452 & \second{0.1277} & 0.8372 & 0.6398 & 0.9212\\
 & + Level Reliability & \second{0.2494} & \second{0.0414} & 0.1279 & 0.8373 & 0.6395 & 0.9213\\
 & + Conf. evidence & \best{0.2502} & 0.1006 & 0.1405 & \best{0.8532} & \best{0.6628} & \best{0.9344}\\
 & \textbf{Full DirEAG} & \ours{\best{0.2502}} & \ours{\best{0.0387}} & \ours{\best{0.1245}} & \ours{\second{0.8513}} & \ours{\second{0.6547}} & \ours{\second{0.9337}}\\
\midrule
\multicolumn{8}{c}{\textbf{Model: Gemma-2-9B-it}}\\
\midrule
GSM8K & Count-only + Cal. & \best{0.8840} & \second{0.0496} & \best{0.0791} & 0.8208 & 0.9592 & \second{0.4813}\\
 & + Level Reliability & \best{0.8840} & 0.0501 & \second{0.0792} & \second{0.8210} & \second{0.9593} & \best{0.4821}\\
 & + Conf. evidence & \best{0.8840} & 0.3841 & 0.2281 & \best{0.8342} & \best{0.9660} & 0.4656\\
 & \textbf{Full DirEAG} & \ours{\best{0.8840}} & \ours{\best{0.0292}} & \ours{0.0811} & \ours{\best{0.8342}} & \ours{\best{0.9660}} & \ours{0.4656}\\
\addlinespace[1pt]
SVAMP & Count-only + Cal. & \best{0.8230} & \best{0.0306} & \second{0.1117} & 0.7229 & \best{0.8938} & 0.5189\\
 & + Level Reliability & \best{0.8230} & \second{0.0318} & 0.1120 & 0.7228 & \best{0.8938} & 0.5160\\
 & + Conf. evidence & \best{0.8230} & 0.3315 & 0.2257 & \best{0.7306} & 0.8904 & \best{0.5364}\\
 & \textbf{Full DirEAG} & \ours{\best{0.8230}} & \ours{0.0418} & \ours{\best{0.1112}} & \ours{\second{0.7293}} & \ours{\second{0.8908}} & \ours{\second{0.5335}}\\
\addlinespace[1pt]
GSM-Hard & Count-only + Cal. & \second{0.5027} & \second{0.0437} & 0.1834 & 0.7837 & 0.7253 & 0.7951\\
 & + Level Reliability & \second{0.5027} & 0.0457 & 0.1837 & 0.7834 & 0.7252 & 0.7952\\
 & + Conf. evidence & \best{0.5042} & 0.0892 & \second{0.1786} & \best{0.8285} & \best{0.8016} & \best{0.8172}\\
 & \textbf{Full DirEAG} & \ours{\best{0.5042}} & \ours{\best{0.0432}} & \ours{\best{0.1679}} & \ours{\second{0.8274}} & \ours{\second{0.8010}} & \ours{\second{0.8130}}\\
\end{longtable}
\endgroup

\paragraph{Analysis.}
The ablation separates candidate-level evidence aggregation from final probability-scale correction. Count-only + Cal. is already a strong baseline, confirming that repeated answer identity is an important signal for mathematical reasoning. Adding steering-level reliability alone produces only minor changes, suggesting that prompt-level weights are not the main source of improvement. In contrast, adding reported confidence as evidence can alter candidate scoring and sometimes the selected answer, but its raw posterior mass is poorly calibrated, leading to high ECE and Brier score in several settings.

Full DirEAG preserves the candidate-level effects of confidence evidence while applying final binary calibration to correct the selected-answer probability scale. Since the binary calibrator is monotonic and applied only after answer selection, it does not change which candidate is selected. Thus, accuracy changes relative to count-only variants come from the evidence aggregation stage, whereas ECE and Brier improvements mainly reflect correction of the final probability scale.

\subsection{Diagnostic Analysis of Learned Uncertainty}

The main results show that DirEAG improves calibration metrics, but a remaining question is whether the learned confidence scores contain instance-level information or mainly recover the average success rate of each model--dataset pair. To test this, we compare DirEAG with a constant base-rate diagnostic. For each model--dataset pair, the diagnostic assigns every instance the same confidence, equal to DirEAG's empirical accuracy on that pair:
\begin{equation}
\hat c_i^{\mathrm{base}}
=
\bar z
=
\frac{1}{n}\sum_{i=1}^{n}\mathbf{1}[\hat a_i=y_i].
\end{equation}
This diagnostic is intentionally strong but not deployable, because it uses the evaluated set's empirical correctness rate. Among constant predictors, it is optimal for squared error; however, it has no ability to rank easier and harder instances, so its AUROC is $0.5$ when both correct and incorrect examples are present.

\begin{figure}[t]
\centering
\begin{minipage}[t]{0.49\textwidth}
\centering
\includegraphics[width=\textwidth]{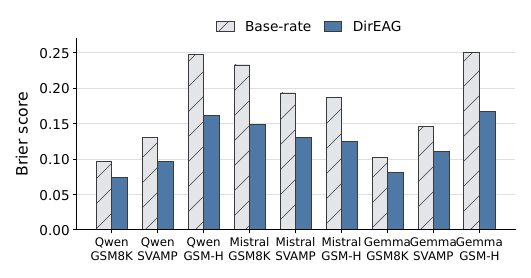}
\small (a) Brier score
\end{minipage}\hfill
\begin{minipage}[t]{0.49\textwidth}
\centering
\includegraphics[width=\textwidth]{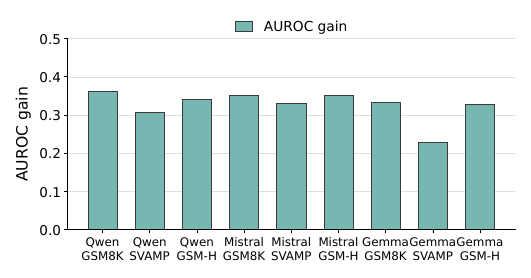}
\small (b) AUROC gain
\end{minipage}
\caption{Instance-level diagnostic against a constant base-rate predictor. Panel (a) compares Brier scores directly, where lower is better; color and hatching distinguish Base-rate from DirEAG. Panel (b) reports AUROC improvement over the base-rate predictor, whose AUROC is $0.5$.}
\label{fig:instance-level-diagnostic}
\end{figure}

Figure~\ref{fig:instance-level-diagnostic} shows that DirEAG obtains lower Brier score than the constant diagnostic in all nine model--dataset settings, with absolute reductions ranging from 0.021 to 0.086. More importantly, DirEAG achieves AUROC gains of 0.229 to 0.362 over the constant predictor. These gains indicate that the learned aggregation is not only matching marginal accuracy; it also uses the pattern of answer-confidence observations to assign different uncertainty levels to different problem instances.

\begin{figure}[t]
\centering
\includegraphics[width=.95\textwidth]{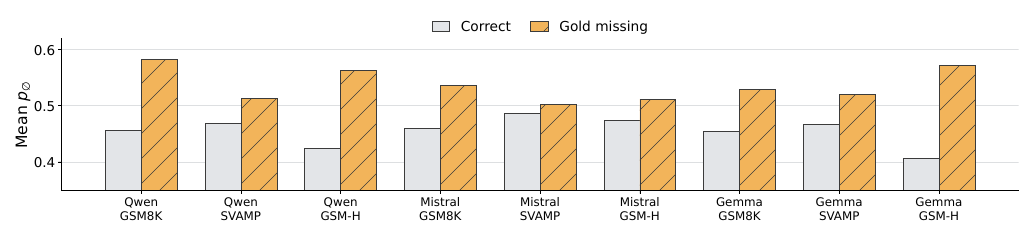}
\caption{Mean null-state probability for correct predictions and gold-absent candidate sets.}
\label{fig:null-state-diagnostic}
\end{figure}

We further examine the learned probability assigned to the null state. As shown in Fig.~\ref{fig:null-state-diagnostic}, this probability is consistently higher when the gold answer is missing from the generated answers than when the final prediction is correct. This suggests that the null state responds to cases where the observed answers do not contain the correct solution.

\section{Conclusion}

We presented DirEAG, a learned aggregation method for verbalized confidence in mathematical reasoning. Across GSM8K, SVAMP, and GSM-Hard, DirEAG improves calibration and remains competitive in answer selection compared with direct verbalized-confidence and heuristic confidence-steering baselines. The ablation indicates that using reported confidence and calibrating the final selected-answer score play complementary roles. Overall, the results support treating verbalized confidence as informative but scale-dependent evidence, rather than as a finished probability estimate.

\paragraph{Limitations.}
This study focuses on numerical mathematical reasoning, where final-answer correctness can be evaluated by exact numeric matching. This controlled setting does not cover open-ended generation, dialogue, proof-oriented reasoning, multimodal tasks, or domains where answer equivalence requires semantic judgment; extending DirEAG to such settings would require task-specific verification and candidate-state definitions. Our experiments also use moderate-scale open-weight models rather than the newest frontier models, because very strong models may produce too few errors on standard arithmetic benchmarks for failure-aware calibration analysis. Future work should evaluate stronger models on harder benchmarks where both correct and incorrect predictions remain sufficiently represented.

\bibliographystyle{splncs04}
\bibliography{reference}

\section*{Appendix}

\subsection*{Experimental Alignment and Ablation Variants}

Each black-box method uses five observations per problem: DirEAG, mean confidence, and SteerConf use five steering prompts; sample-based baselines use five vanilla chain-of-thought samples; and Top-$K$ asks for five candidate answers in one response. For Mistral, we observe that the Top-$K$ ranked list can be misaligned with the subsequent reasoning trace: in some cases, the explanation derives the correct numerical answer, while the highest-confidence listed candidate is different. We therefore treat Top-$K$ as a direct multi-candidate elicitation baseline whose reliability depends on faithful ranked-list generation.

The ablation reuses the same five answer-confidence observations and changes only aggregation and calibration. Let $(a_\ell,q_\ell)$ be the answer and confidence from level $\ell$, $C$ the candidate set, and $w_\ell$ the learned level weight. Define
\begin{equation}
\rho_\ell=\sigma(b_\ell),
\qquad
\tilde q_\ell=\sigma\left(b_\ell+s\,\mathrm{logit}(q_\ell)\right),\quad s>0,
\qquad
\hat y=\arg\max_{c\in C}\alpha_c.
\end{equation}
The variants differ as follows, where $\mathrm{Cal}(\cdot)$ denotes final binary calibration.

\paragraph{Count-only + Cal.}
\begin{equation}
\begin{array}{rcl}
\alpha_c&=&\alpha_0+\sum_{\ell=1}^{5}w_\ell\mathbf{1}[a_\ell=c],\\
\alpha_{\emptyset}&=&\alpha_0+\beta_0,\qquad
\hat p=\mathrm{Cal}\!\left(\frac{\alpha_{\hat y}}{\sum_{c'\in C}\alpha_{c'}+\alpha_{\emptyset}}\right).
\end{array}
\end{equation}

\paragraph{+ Level Reliability.}
\begin{equation}
\begin{array}{rcl}
\alpha_c&=&\alpha_0+\sum_{\ell=1}^{5}w_\ell\rho_\ell\mathbf{1}[a_\ell=c],\\
\alpha_{\emptyset}&=&\alpha_0+\beta_0+\gamma\sum_{\ell=1}^{5}w_\ell(1-\rho_\ell),\\
\hat p&=&\mathrm{Cal}\!\left(\frac{\alpha_{\hat y}}{\sum_{c'\in C}\alpha_{c'}+\alpha_{\emptyset}}\right).
\end{array}
\end{equation}

\paragraph{+ Conf. Evidence.}
\begin{equation}
\begin{array}{rcl}
\alpha_c&=&\alpha_0+\sum_{\ell=1}^{5}w_\ell\tilde q_\ell\mathbf{1}[a_\ell=c],\\
\alpha_{\emptyset}&=&\alpha_0+\beta_0+\gamma\sum_{\ell=1}^{5}w_\ell(1-\tilde q_\ell),\\
\hat p&=&\frac{\alpha_{\hat y}}{\sum_{c'\in C}\alpha_{c'}+\alpha_{\emptyset}}.
\end{array}
\end{equation}

\paragraph{Full DirEAG.}
\begin{equation}
\begin{array}{rcl}
\alpha_c&=&\alpha_0+\sum_{\ell=1}^{5}w_\ell\tilde q_\ell\mathbf{1}[a_\ell=c],\\
\alpha_{\emptyset}&=&\alpha_0+\beta_0+\gamma\sum_{\ell=1}^{5}w_\ell(1-\tilde q_\ell),\\
\hat p&=&\mathrm{Cal}\!\left(\frac{\alpha_{\hat y}}{\sum_{c'\in C}\alpha_{c'}+\alpha_{\emptyset}}\right).
\end{array}
\end{equation}
\end{document}